\pdfoutput=1

\documentclass[11pt]{article}

\usepackage{acl}

\usepackage{times}
\usepackage{latexsym}
\usepackage{amssymb}
\usepackage{amsmath}
\usepackage{enumitem}
\usepackage{multirow}
\usepackage{booktabs}
\usepackage{graphicx}
\usepackage{textcomp}
\usepackage{float}
\usepackage{xcolor}
\usepackage{bibentry}
\usepackage{placeins}
\usepackage{hyperref}

\usepackage[T1]{fontenc}

\usepackage[utf8]{inputenc}

\usepackage{microtype}
\usepackage{graphicx}

\newcommand{\secref}[1]{\S\ref{#1}}
\newcommand{\figref}[1]{Fig.~\ref{#1}}
\newcommand{\tabref}[1]{Table~\ref{#1}}
\title{Few-Shot Table-to-Text Generation with Prefix-Controlled Generator}

\author{Yutao Luo , Menghua Lu , Gongshen Liu\thanks{~~Corresponding author.} , Shilin Wang \\
  Shanghai Jiao Tong University\\
  \texttt{\{luoyt1996, 610228633, lgshen, wsl\}@sjtu.edu.cn}\\}

\begin{document}
\maketitle
\begin{abstract}
Neural table-to-text generation approaches are data-hungry, limiting their adaptation for low-resource real-world applications. Previous works mostly resort to Pre-trained Language Models (PLMs) to generate fluent summaries of a table. However, they often contain hallucinated contents due to the uncontrolled nature of PLMs. Moreover, the topological differences between tables and sequences are rarely studied. Last but not least, fine-tuning on PLMs with a handful of instances may lead to over-fitting and catastrophic forgetting. To alleviate these problems, we propose a prompt-based approach, Prefix-Controlled Generator (i.e., PCG), for few-shot table-to-text generation. We prepend a task-specific prefix for a PLM to make the table structure better fit the pre-trained input. In addition, we generate an input-specific prefix to control the factual contents and word order of the generated text. Both automatic and human evaluations on different domains (humans, books and songs) of the Wikibio dataset show substantial improvements over baseline approaches.
\end{abstract}

\section{Introduction}
\label{intro}
Table-to-text generation is a significant branch of Natural Language Generation (NLG), aiming at generating descriptive text given an input table. There is a wide range of application scenarios for automatic table-to-text generation, such as sport news generation \cite{rotowire}, story generation \cite{ldy}, weather forecasting report \cite{weathergov}, and open-domain question answering \cite{iclr21ottqa}.

Recent years have witnessed the great development of pre-trained language models (PLMs) \cite{bert, gpt2, bart}, which achieve state-of-the-art performance on many text generation tasks, such as neural machine translation, document summarization, etc. Unlike these tasks, table-to-text generation faces the lack of labeled data. Due to the development of data science, many statistical tables are generated in our daily life, but they scarcely have corresponding natural language descriptions, which limits the real-world application of data-hungry pre-trained models. To address this problem, researchers investigate workarounds in the few-shot setting. \citet{ChenACL20}, \citet{GongCOLING20} and \citet{su2021few} leverage pre-trained linguistic knowledge of neural language models, then fine-tune them in target domains with limited labeled data. This ``\textit{pre-train and fine-tune}'' paradigm performs relatively well in generating descriptive text from tables. Recently, another paradigm named ``\textit{pre-train and prompt}'' has been proposed in order to adapt PLMs to downstream tasks without fine-tuning, which is more suitable for few-shot and zero-shot scenarios. \citet{li2021prefix} prepends prompt tokens to 
adapt table-to-text generation to sequential generation task, and freezes PLMs' weights to fully leverage their prior knowledge learned in the pre-training stage.

\begin{figure*}[t]
\centering
\includegraphics[width=\textwidth]{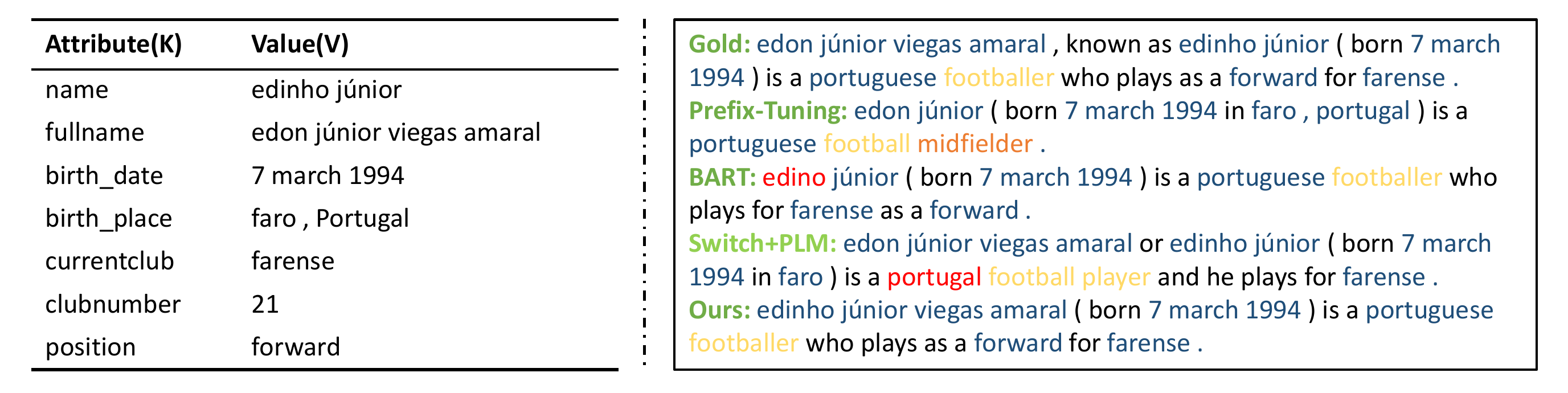} 
\caption{An example from \textit{Wikibio Humans} domain and the generated descriptions via various approaches. Words in blue, red, orange and yellow indicate factual contents, wrong generation, hallucinated contents and inferred contents respectively. BART represents BART-large \cite{bart}. Switch+PLM represents \citet{ChenACL20}'s approach with BART-large.}
\label{fig1}
\end{figure*}

Despite their contributions, however, two main challenges for table-to-text generation remain to be explored, namely (1) the \textbf{topological structure difference} between tables and sequential inputs and (2) model's ability to \textbf{select and rearrange factual content} from tables.

In order to address the aforementioned problems, we follow the ``\textit{pre-train and prompt}'' paradigm and propose Prefix-Controlled Generator (i.e., PCG), an end-to-end generation framework along with two kinds of prefix tokens. Specifically, we prepend a task-specific (i.e., static) prompt and an input-specific (i.e., dynamic) prompt to the tabular input. The task-specific prompt aims to bridge the topological structure gap between a table and a word sequence, while the input-specific prompt aims to plan the factual content and the slot order of a table. Both prefixes are optimized during the training phase with the PLM remaining frozen, making our approach parameter-efficient -- we only save one copy of the PLM while training in three different domains.

We basically follow the idea of prefix-tuning \cite{li2021prefix} to design the task-specific prefix, except for some modifications. Firstly, due to the importance of a proper initialization of prefix tokens, we use task-relevant words (e.g., ``\textit{Summarize the following table:}'', or ``\textit{TL;DR:}'') as the initial prefix to better linearize the tabular input. Secondly, \citet{iclr2022he} proves that the length of prefix tokens and the design of adding additional parameters solely on the attention module are two bottlenecks of prefix-tuning. Inspired by their work, we add Scaled Parallel Adapters \cite{iclr2022he} in parallel with both the attention layer and the feed-forward layer to improve the bottleneck of prefix-tuning.

For the input-specific prefix, we expect it can hint to the model which key-value pairs should be selected and in what order they should be arranged. Therefore, we propose a content planner to select the keys that appear in the gold summary and sort them according to the order of occurrence in the summary. For example, given a table in \figref{fig1}, we expect the content planner to generate a word sequence ``\textit{fullname name birth\_date birth\_place position currentclub}'' that indicates all the keys and their occurrence order whose values appear in the gold summary. The word sequence will be used as hard prompts to feed into the PLM.

We evaluate our model on multi-domain table-to-text dataset \cite{ChenACL20}. We show that our model outperforms previous state-of-the-art methods on both automatic evaluation metrics (\secref{5-res}) and human evaluation metrics (\secref{5-hum}). We also conduct ablation studies to verify the effectiveness of the two kinds of prefixes (\secref{5-abl}).

In a nutshell, our contributions are as follows:

\begin{enumerate}
\item [1. ] We propose a Prompt-Controlled Generator that attends to the task-specific prefix to bridge the topological structure gap between tables and sequences, and the input-specific prefix to select factual contents from the tables and reorder them.
\item [2. ] We propose a simple yet effective content planner to generate the input-specific prefix as the hard prompt of the PLM.
\item [3. ] We conduct experiments on different domains of the Wikibio dataset to prove the effectiveness of our approach.
\end{enumerate}

\section{Related Work}

\subsection{Few-shot Table-to-text generation}
Table-to-text generation has aroused much interest in recent years. Most of the existing studies resort to the end-to-end framework to generate fluent and faithful natural language descriptions given tables. \citet{ma2019key} firstly studied table-to-text generation under the low-resource constraint, and separated the generation process into two stages -- key fact prediction and surface realization. With the advances of PLMs, many researchers fine-tune pre-trained GPT-2 \cite{gpt2} or BART \cite{bart} to augment the scarce training data, which can better assist few-shot table-to-text generation. \citet{ChenACL20} used copy mechanism \cite{pgen} to improve the fidelity of sentences generated by GPT-2 by choosing to copy words from tabular input. \citet{GongCOLING20} adopted a unified GPT-2 model for table structure reconstruction and generation. \citet{amg} proposed a token-level attention and a slot-level attention to exploit natural linguistic and table structural information. All these works utilized tabular input for free text generation, neglecting the importance of content planning for text fidelity. \citet{su2021few} introduced an information retrieval (IR) system to select prototype sentences similar to the gold summary from large unlabeled parallel corpus, then use them as the auxiliary content plan for tabular input to generate natural language description. However, the IR system might see all gold summaries in the Wikipedia corpus, which violates the true few-shot setting. Different from the above studies, we focus on how to select factual contents via content planning, introducing a slot-aligned content planner.

\subsection{Prompt Tuning for Generation}
Prompt tuning is a nascent approach for natural language generation (NLG), first proposed by GPT-3 \cite{gpt3}, introducing in-context learning for few-shot domain adaptation. Prefix-tuning \cite{li2021prefix} prepended a sequence of continuous vectors to all examples of the downstream tasks. These vectors, which are adjusted as additional key-value pairs, steer the frozen PLMs by augmenting the left context at every Transformer layer. \citet{iclr2022he} classified prefix-tuning as a parameter-efficient tuning approach similar to adapter \cite{adapter} and made improvements on its bottlenecks. \citet{controlprefix} extended prefix-tuning to the input-specific prefix (e.g., topic of the datapoint, target output length) to have a finer-grained control for downstream generation tasks. Different from their work, we use the input-specific prompt not to guide the generated text in a certain style, but to improve the fidelity of generated text and the correctness of word order via content planning.

\subsection{Controllable Text Generation}
Controllable text generation (CTG) is a supplementary field for prompt-based generation, aiming at incorporating guidance signals into generative models. Control signals include text style \cite{CTRL}, grammar \cite{styleptb}, length \cite{kikuchi}, etc. Recent CTG approaches involve generative adversarial networks \cite{seqgan}, refactoring a PLM \cite{cocon}, fine-tuning adapted modules \cite{auxadapter}, prompt learning \cite{emnlpfindings21yu} and diffusion model \cite{diffusionlm}. However, these approaches requires large amount of training data, which does not match our few-shot setting. For table-to-text generation, \citet{plangen} proposed a content planner to assist data-to-text generation, which inspired us to pre-plan the order and occurrence of the tabular input for improving the controllability of the generated text.

\section{Problem Definition}
Given a table $T$ with $n$ key-value pairs $\{\mathbf{K}_i:\mathbf{V}_i\}_{i=1}^n$, where $\mathbf{K}_i=\{k_1^i,k_2^i,...,k_m^i\}$ and $\mathbf{V}_i=\{v_1^i,v_2^i,...,v_{m'}^i\}$ refer to the key and the value of the $i$-th table slot respectively, we aim to generate a fluent and faithful natural language description of the table in a low-resource constraint. Note that $\mathbf{K}_i$ and $\mathbf{V}_i$ represent sequences of $m$ and $m'$ words respectively.

\begin{figure*}[t]
\centering
\includegraphics[width=\textwidth]{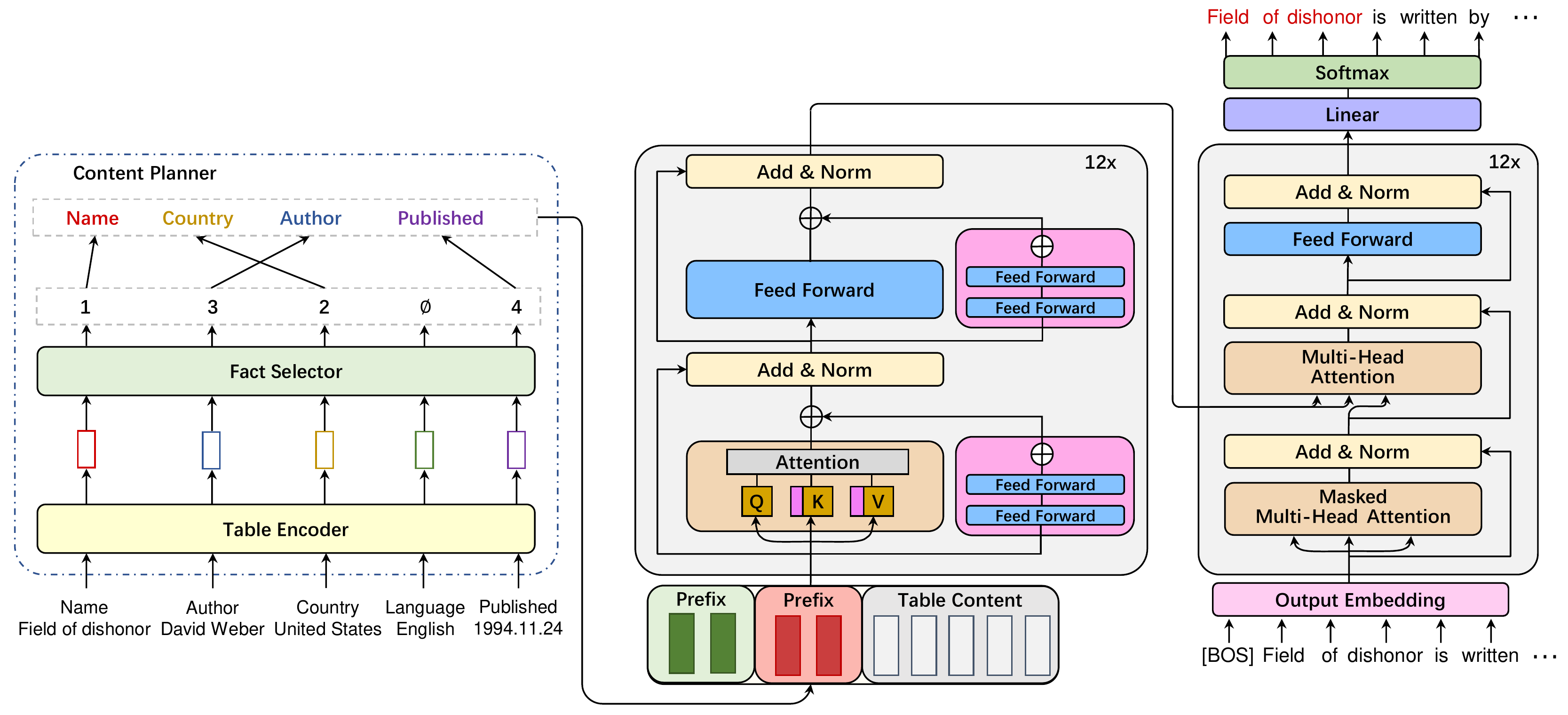} 
\caption{The overall architecture of the proposed method, which can be divided into Content Planner and Prompt-Controlled Generator. Tokens in red and yellow indicate these words are consistent with the value corresponding to the key ``\textit{name}'' and ``\textit{country}'' respectively.}
\label{fig2}
\end{figure*}

\section{Methodology}
We first provide intuition of using a task-specific prefix and an input-specific prefix for few-shot table-to-text generation (\secref{intuition}). \figref{fig2} depicts the overall architecture of our method. As shown in the figure, given the input table, the content planner selects the factual contents and reorders them to form a dynamic prompt (\secref{cp}). After that, a static prompt is designed and fed to the generative PLM along with the dynamic prompt (\secref{pcg}).

\subsection{Intuition}\label{intuition}
The intuition of introducing prompt to few-shot table-to-text generation is that prompt-tuning effectively solves the catastrophic forgetting problem. Since table-to-text generation requires the language understanding ability of the table content, we hope to fine-tune downstream tasks while retaining the prior knowledge of PLMs, which is exactly what prompt-tuning does. Unlike model fine-tuning, which might be over-parameterized, prompt tuning only adjusts a few parameters and is less prone to over-fitting.

Observing \citet{GongCOLING20}'s experimental results, we find that table format transformation plays a vital role in improving the generation process, so we focus on bridging the topological structure gap between tables and sequential inputs. The first thought is that we can flatten a table into a word sequence using template \cite{GongCOLING20}. For example, given a table shown in \figref{fig1}, we serialize the key-value pair \{\textit{name: edinho júnior}\} as ``\textit{name is edinho júnior;}'', then concatenate all key-value pairs to form a sentence, that is, ``\textit{name is edinho júnior; fullname is edon júnior viegas amaral; birth\_date is 7 march, 1994; ...}''. Considering that the template-generated sentence is still somewhat different from the pre-training input, we want to find a way that adapts it to a natural sentence. Intuitively, we can add some prompt tokens like ``\textit{summarize the following table:}'' to make the template-generated sentence an incidental component of the whole input. In this way, ``\textit{summarize the following table}'' becomes the major component of the sentence, which is more similar to the sequential form of the pre-training input. In addition, many language models now have prefix LM pre-training tasks, which makes our sentences more consistent with the pre-training input. 

We also seek to minimize the generated hallucinated content. Considering that some table slots are redundant, we intuitively want to hint the model what are the factual contents. To be consistent with the above table linearization approach, we follow the idea of controllable generation, providing a hard prompt as the guidance signal for each example to control both the table content to be selected and the word order.

\subsection{Content Planner}\label{cp}
Content Planner aims to generate input-specific prompts that guide the generation process in terms of factual contents and words order, which is shown in \figref{fig2} (left). Content Planner contains two modules, namely Table Encoder and Fact Selector. Since we study table-to-text generation under a strict few-shot constraint, we strive for simplicity of Content Planner. Therefore, we use a bi-directional LSTM and a linear-chain Conditional Random Fields (CRF) \cite{crf} for Table Encoder and Fact Selector respectively, which are learned given a handful of training instances.

Table Encoder takes all key-value pairs $\{\mathbf{K}_i:\mathbf{V}_i\}_{i=1}^n$ from table $T$ as the input, and produces a hidden representation $\mathbf{h}_i\in \mathbb{R}^{d_e}$ for each table slot, where $d_e$ is the hidden dimension. Specifically, for each table slot that contains a key-value pair $\{\mathbf{K}_i:\mathbf{V}_i\}$, we embed $\mathbf{K}_i$ and $\mathbf{V}_i$ by:
\begin{equation}
    \mathbf{e}_i = \lambda\frac{1}{m}\sum_{j=1}^{m}\mathbf{E}(k_j^i) + (1-\lambda)\frac{1}{m'} \sum_{j=1}^{m'}\mathbf{E}(v_j^i) ,
\end{equation}
where $\mathbf{e}_i$ denotes the embedding of the $i$-th slot, $\mathbf{E}$ is the embedding lookup table and $\lambda$ is a hyper-parameter that controls the ratio of key embedding and value embedding in $\mathbf{e}_i$. We use pre-trained Roberta \cite{roberta} embedding to initialize $\mathbf{E}$. After that, we feed all embeddings $\{\mathbf{e}_1,\mathbf{e}_2,...,\mathbf{e}_n\}$ to the BiLSTM encoder to obtain $\{\overrightarrow{\mathbf{h}_1},\overrightarrow{\mathbf{h}_2},...,\overrightarrow{\mathbf{h}_n
}\}$ and $\{\overleftarrow{\mathbf{h}_1},\overleftarrow{\mathbf{h}_2},...,\overleftarrow{\mathbf{h}_n}\}$ in the left-to-right and right-to-left directions respectively. The calculation of each direction uses a distinct set of parameters. 
The final hidden states $\mathbf{h}_i$ can be obtained by:
\begin{equation}
    \mathbf{h}_i = [\overrightarrow{\mathbf{h}_i};\overleftarrow{\mathbf{h}_i}].
\end{equation}

Fact Selector selects key-value pairs that occur in the ground-true table summary, and rearranges them according to the order of occurrence in the summary. In practice, we use a standard CRF layer with a feed-forward layer as our Fact Selector to compute the global optimal sequence. On top of the hidden states $\mathbf{H}_c=\{\mathbf{h}_1,\mathbf{h}_2,...,\mathbf{h}_n\}$, the probability distributions of the label sequence $\mathbf{y}=\{l_1,l_2,...,l_n\}$ is computed by:

\begin{small} 
\begin{equation}
    P(\mathbf{y}\vert \mathbf{H}_c) = \frac{exp(\sum_{i}(\mathbf{W}_{CRF}^{l_i}\mathbf{h}_i+\mathbf{M}_{l_{i-1},l_i}))}{\sum_{\mathbf{y'}} exp(\sum_{i}(\mathbf{W}_{CRF}^{l'_i}\mathbf{h}_i+\mathbf{M}_{l'_{i-1},l'_i}))}.
\end{equation}
\end{small}\\
Here $\mathbf{y'}$ represents an arbitrary label sequence, $\mathbf{W}_{CRF}^{l_i}$ denotes the parameters specific to $l_i$, and $\mathbf{M}_{l_{i-1},l_i}$ denotes the transition score from $l_{i-1}$ to $l_i$. The learning objective is defined as:
\begin{equation}
    \mathcal{L}_{CRF}=-logP(\mathbf{y}\vert \mathbf{H}_c).
\end{equation}
Content Planner is trained independently with Prompt-Controlled Generator. The labeled key-value pair order is extracted from the ground-true summary by finding keys\footnote{Some keys such as ``nationality'' are fuzzy-matched.} and sorting them according to their positions. During inference, we use first-order Viterbi algorithm to decode the best label sequence $\mathbf{\widetilde{y}}=argmax_{\mathbf{y'}}P(\mathbf{y'}\vert \mathbf{H}_c)$. Take \figref{fig2} as an example, Content Planner generates a label sequence ``1,3,2,$\emptyset$,4''. The first label ``1'' denotes ``\textit{Name}'' should appear in the front of the content plan, while the fourth label ``$\emptyset$'' denotes ``\textit{Language}'' does not occur in the gold summary. According to the label sequence, we rearrange all keys to form a content plan $\mathbf{c}$, which in \figref{fig2} is ``\textit{Name Country Author Published}''.

\begin{table*}[t]
\renewcommand\arraystretch{1.05}
\centering
\resizebox{0.98\textwidth}{!}{
\begin{tabular}{ccccccccccccc}
\toprule
 Domain & \multicolumn{4}{c}{Humans} & \multicolumn{4}{c}{Books} & \multicolumn{4}{c}{Songs}\\ 
\cmidrule(lr){1-1} \cmidrule(lr){2-5} \cmidrule(lr){6-9} \cmidrule(lr){10-13}
Training set size & 50 & 100 & 200 & 500 & 50 & 100 & 200 & 500 & 50 & 100 & 200 & 500    \\
\midrule
\textbf{Switch+GPT-2(R)} &  25.7&29.5&36.1& 41.7  &    34.3&36.2&37.9&40.3   &    36.1&37.2&39.4&42.2   \\
\textbf{TableGPT(R)}    &  29.8&34.5&40.6&45.6   &    35.1&\textbf{37.3}&38.5&41.6    &    36.7&37.8&39.3&42.3\\
\textbf{Bart-large}    & 37.6&39.3&41.2&44.3   &    34.2&37.1&\textbf{39.8}&42.9    &    37.7&38.9&40.1&43.9\\
\textbf{AMG(R)}        & -&-&-&49.0   &    -&-&-&43.9    &    -&-&-&\textbf{45.1}\\
\midrule
\textbf{Hard-prompt+GPT-2} & 22.8&28.1& 29.7& 30.8&     25.8&27.9& 28.8 &32.1&       26.6&30.0&30.1&32.1 \\
\textbf{Prefix-Tuning+GPT-2} & 25.6&30.3& 33.4&37.3&     34.9&36.2& 36.3 &37.3&       32.5&33.0&35.1&36.1 \\
\textbf{Prefix-Tuning+T5}  &34.5&39.9& 41.6&44.1&   35.5& \textbf{37.3}& 39.6 &41.2     &37.5&38.5&40.0&41.1   \\
\textbf{Switch+BART(PT)}     &  36.8&41.8&44.0&48.1  &    33.6&35.0&38.3&43.4   &    \textbf{40.9}&\textbf{41.7}&42.1&43.2   \\
\midrule
\textbf{Ours}    & \textbf{39.9}&\textbf{43.3}&\textbf{45.8}&\textbf{49.4}  &  \textbf{36.6}& 36.9 & 39.0&\textbf{45.6}   &   38.0&\textbf{41.7}&\textbf{42.5}&44.5   \\
\bottomrule
\end{tabular}}
\caption{BLEU results on three domains of the Wikibio test set. Each \textbf{(R)} is reported by the related paper.}
\label{tab:bleu_eval}
\end{table*}

\begin{table*}[t]
\renewcommand\arraystretch{1.05}
\centering
\resizebox{0.98\textwidth}{!}{
\begin{tabular}{ccccccccccccc}
\toprule
 Domain & \multicolumn{4}{c}{Humans} & \multicolumn{4}{c}{Books} & \multicolumn{4}{c}{Songs}\\ 
\cmidrule(lr){1-1} \cmidrule(lr){2-5} \cmidrule(lr){6-9} \cmidrule(lr){10-13}
Training set size & 50 & 100 & 200 & 500 & 50 & 100 & 200 & 500 & 50 & 100 & 200 & 500    \\
\midrule
\textbf{Switch+GPT-2(R)}    & 30.6&34.6&40.5&45.6& 42.7&42.8&43.4&44.9 & 40.2&41.7&44.0&44.8   \\
\textbf{Bart-large(R)}    & 37.8&41.4&47.4&45.5    &  41.7&43.4&43.7&48.1    &    41.7&42.4&44.1&46.0   \\
\textbf{AMG(R)}    & 43.6&47.7&50.1&\textbf{51.9}    &   43.4&46.0&\textbf{47.5}&48.6   & 42.0&43.3&45.9&\textbf{46.9} \\
\midrule
\textbf{Prefix-Tuning+GPT-2} & 32.7&35.9&36.6&38.7&     29.8&31.8&31.7&32.7&    31.7&33.3&32.3&31.5    \\
\textbf{Prefix-Tuning+T5}  &  39.3&40.6&41.8&42.1&    32.8&34.8&36.0&36.8&     34.4&36.1&36.0&34.6    \\

\textbf{Switch+BART(PT)}     &  35.2&41.7&45.1& 50.5  &    33.0&37.2&41.2&46.4   &    36.7&39.4&42.0&45.9   \\
\midrule
\textbf{Ours} & \textbf{46.7}&\textbf{48.3}&\textbf{50.4}&51.8   &   \textbf{46.3}&\textbf{46.2}&\textbf{47.5}&\textbf{49.3}    &   \textbf{44.8}& \textbf{45.7}&\textbf{46.9}&46.0   \\
\bottomrule
\end{tabular}}
\caption{PARENT-F results on three domains of the Wikibio test set. All \textbf{(R)} are reported by \citet{amg}.}
\label{tab:parent_eval}
\end{table*}

\subsection{Prompt-Controlled Generator}\label{pcg}

Prompt-Controlled Generator aims to generate fluent and faithful descriptions given the tabular input and the content plan. Our approach is model-agnostic, thus the generator could be any pre-trained generation model. Here we use BART-large \cite{bart} as the basic generator for their best overall performances, and propose two kinds of prefixes that are prepended to the input of BART encoder, namely task-specific prompt $\mathbf{p}_s$ and input-specific prompt $\mathbf{c}$. The latter (i.e., content plan) serves as the guiding signal of Prompt-Controlled Generator.

The task-specific prompt is designed to bridge the topological structure gap between tables and sequences. A first thought is that we can linearize the table via template (see \secref{intuition}), then prepend discrete prompt words ``\textit{summarize the following table:}'' to the template-generated sequence to make the tabular input more consistent with the pre-training input. Nevertheless, discrete optimization needs enormous computing power and human crafts. Instead of using discrete prompt, we follow prefix-tuning \cite{li2021prefix} to optimize a sequence of continuous prefix tokens while keeping the PLM frozen. However, the prefix length and acting on the attention layer bound the presentation ability of the prefix \cite{iclr2022he}. Considering these bottlenecks, we additionally parallel two Scaled Parallel Adapters to the attention layer and the feed-forward layer respectively, then perform scaled addition for these Adapters.

Next, we will introduce our modifications to BART encoder. Let us denote the template-generated sentence as $\mathbf{s}=\{s_1,s_2,...,s_L\}$ and content plan as $\mathbf{c}=\{c_1,c_2,...,c_{L_c}\}$, where $L$ and $L_c$ are the lengths. The prefix length is denoted by $L_p$. We concatenate the content plan and the template-generated sentence (denoted by $[\mathbf{c}\colon \mathbf{s}]$ where $[\cdot \colon \cdot]$ is the concatenation operator) to feed into BART encoder. In the multi-head self-attention layer, we first compute the queries $\mathbf{Q}\in \mathbb{R}^{(L+L_c)\times d}$, keys $\mathbf{K}\in \mathbb{R}^{(L+L_c)\times d}$ and values $\mathbf{V}\in \mathbb{R}^{(L+L_c)\times d}$ via Equation \eqref{eq1}:
\begin{equation}\label{eq1}
    \mathbf{Q}=\mathbf{x}\mathbf{W}_q, \mathbf{K}=\mathbf{x}\mathbf{W}_k, \mathbf{V}=\mathbf{x}\mathbf{W}_v,
\end{equation}
where $d$ denotes the hidden dimension of BART, $\mathbf{W}_q$, $\mathbf{W}_k$ and $\mathbf{W}_v$ are trainable parameters. $\mathbf{x}$ denotes $\mathbf{E_b}([\mathbf{c}\colon \mathbf{s}])$ when the first layer is being computed, the output of the previous BART layer otherwise. $\mathbf{E_b}$ denotes the embedding lookup table of BART. Then the attention score is computed via Equation \eqref{eq2}:
\begin{equation}\label{eq2}
    head = [head^1\colon head^2\colon...\colon head^{N_h}],
\end{equation}
where $N_h$ denotes the number of heads. $head^i$ is computed via Equation \eqref{eq3}:

\begin{small}
\begin{equation}\label{eq3}
\begin{split}
	head^i&=Attn(\mathbf{Q}^i,\mathbf{K}^i,\mathbf{V}^i)\\
	&=softmax(\frac{\mathbf{Q}^i[\mathbf{P}_k^i\colon\mathbf{K}^i]^T}{\sqrt{d_k}})[\mathbf{P}_v^i\colon\mathbf{V}^i],
\end{split}
\end{equation}
\end{small}\\
where $d_k=\frac{d}{N_h}$ denotes the hidden dimension of each head, and $\mathbf{P}_k\in \mathbb{R}^{L_p\times d}, \mathbf{P}_v\in \mathbb{R}^{L_p\times d}$ denote two sets of prefix vectors. $\mathbf{Q}^i\in \mathbb{R}^{(L+L_c)\times d_k}$, $\mathbf{K}^i\in \mathbb{R}^{(L+L_c)\times d_k}$, $\mathbf{V}^i\in \mathbb{R}^{(L+L_c)\times d_k}$, $\mathbf{P}_k^i\in \mathbb{R}^{(L_p)\times d_k}$ and $\mathbf{P}_v^i\in \mathbb{R}^{(L_p)\times d_k}$ denote a block of $\mathbf{Q}$, $\mathbf{K}$, $\mathbf{V}$, $\mathbf{P}_k$, and $\mathbf{P}_v$ respectively.

In parallel with the multi-head self-attention layer, a Scaled Parallel Adapter is added:
\begin{gather}
    head'=\mathbf{x}+s\cdot ReLU(\mathbf{x}\mathbf{W}_{down})\mathbf{W}_{up},\\
    \mathbf{H}_{attn}=head+head',
\end{gather}
where $\mathbf{W}_{down}\in\mathbb{R}^{d\times r}$ and $\mathbf{W}_{up}\in\mathbb{R}^{r\times d}$ are down-projection and up-projection, $r$ is the bottleneck dimension and $\mathbf{x}$ denotes the same vector as in Equation \eqref{eq1}. $s\geq1$ is a trainable scaling hyper-parameter.
We use $\mathbf{H}_{attn}$ to replace the original attention output to conduct residual connection and layer normalization. Similarly, we insert another Scaled Parallel Adapter in parallel with the Feed Forward layer to enhance its representation:
\begin{equation}
\begin{split}
    \mathbf{o}&=ReLU(\mathbf{x}\mathbf{W}_1+\mathbf{b}_1)\mathbf{W}_2+\mathbf{b}_2\\
    &+s\cdot ReLU(\mathbf{x}\mathbf{W'}_{down})\mathbf{W'}_{up},\\
\end{split}
\end{equation}
where $\mathbf{x}$ and $\mathbf{o}$ denote the input and output of the Feed Forward layer respectively. $\mathbf{W}_1$, $\mathbf{b}_1$, $\mathbf{W}_2$, $\mathbf{b}_2$, $\mathbf{W'}_{down}$, $\mathbf{W'}_{up}$ are trainable parameters.

We conduct residual connection and layer normalization over $\mathbf{o}$ to get the hidden states of BART encoder $\mathbf{H}_{enc}$, then feed $\mathbf{H}_{enc}$ along with decoder input to a normal BART decoder for sentence generation. The decoder input is the right-shifted gold summary in the training phase, and a simple ``[BOS]'' in the inference phase to generate tokens autoregressively. Given the gold summary $\mathbf{g}$, the learning objective is the cross-entropy loss, defined as:
\begin{equation}
    \mathcal{L}_{LM} = -\sum_{i=1}^{\lvert \mathbf{g} \rvert}logP_{dec}(\mathbf{g}_i\vert \mathbf{g}_{<i};\mathbf{H}_{enc}).
\end{equation}

\section{Experiments}
\subsection{Datasets and Hyper-Parameters}
Following \citet{ChenACL20}, we evaluate our method on three different domains (i.e., \textit{Humans}, \textit{Books} and \textit{Songs}) of the Wikibio dataset, denoted as Wiki-Humans, Wiki-Songs and Wiki-Books respectively. 
For all three domains, we conduct experiments in few-shot settings by varying the training set size to 50, 100, 200 and 500. The validation size is set to 1000, and the remaining instances are used for testing, which counts 13587, 5252 and 11879 for humans, books and songs respectively. 

We use BART-large as our basic generator using transformers library \cite{transformers}, which shares 12 layers and 16 heads for both encoder and decoder. We set the hidden and the embedding dimension of Content Planner to 768 (Roberta-base embedding dimension), and the key-value ensemble ratio $\lambda$ is set to 0.7. The learning rates of Content Planner and Prefix-Controlled Generator are set to 2e-4 and 1e-5 respectively, both optimized by AdamW \cite{adamw}. We train our PCG for 200 epochs, with a batch size of 10 on one NVIDIA GeForce RTX 3090 GPU. Prefix length $L_p$ is set to 30, and the bottleneck size of Scaled Parallel Adapter $r$ is set to 512.

\subsection{Baseline Models}
We compare previous state-of-the-art few-shot table-to-text generation approaches, serving as baseline models:
\begin{enumerate}[label=(\roman*)]
    \item \textbf{Switch+PLM} \cite{ChenACL20}: The first work that introduces PLMs to the few-shot NLG task. They propose a switch policy to choose whether to copy words from the table or to generate from GPT-2. We also implement a variant using BART-large to replace GPT-2 and tuning the BART-large model with our task-specific prompt (denoted as \textbf{Switch+BART(PT)}).
    \item \textbf{TableGPT} \cite{GongCOLING20}: A further study based on Switch+PLM that leverages GPT-2's prior knowledge, while enhancing generation fidelity with two auxiliary tasks.
    \item \textbf{AMG} \cite{amg}: A pre-train and fine-tune approach with a multi-grain attention to both tokens and slots, and introduces memory mechanism to back-track the allocation of table slots.
    \item \textbf{BART-large} \cite{bart}: A powerful PLM for conditional generation, which is proved effective in the few-shot scenario \cite{amg}. We fine-tune it on our few-shot datasets to report its performance.
    \item \textbf{Hard-prompt+GPT-2}: Our earlier attempt on the few-shot table-to-text generation task, which uses actual tokens such as "Summarize the following table:" as the prompt words, then feed the transformed tabular input into GPT-2 to fine-tune on few-shot table-to-text generation task.
    \item \textbf{Prefix-Tuning} \cite{li2021prefix}: A novel prompt-based approach that prepends a continuous prefix and freezes the PLMs to retain their prior knowledge. We follow \citet{openprompt}'s implementation, using GPT-2 and T5 \cite{t5} as the base model.
\end{enumerate}
Among above baseline approaches, \textbf{Prefix-Tuning} and \textbf{Switch+BART(PT)}) follow a ``pre-train and prompt-tuning'' paradigm (keep LM's parameters frozen), while \textbf{Hard-prompt+GPT-2} uses prompt for model fine-tuning. All the other baselines are following the standard ``pre-train and fine-tune'' paradigm.

\begin{table}[t]
    \centering
    \resizebox{\columnwidth}{!}{
    \begin{tabular}{ccccccccc}
    \toprule
    \multirow{2}{*}{Model}&\multicolumn{2}{c}{50}&\multicolumn{2}{c}{100}&\multicolumn{2}{c}{200}&\multicolumn{2}{c}{500}\\
    \cmidrule(lr){2-3} \cmidrule(lr){4-5} \cmidrule(lr){6-7} \cmidrule(lr){8-9}
    &acc.&BLEU&acc.&BLEU&acc.&BLEU&acc.&BLEU\\
    \midrule
    Roberta-base&0.32&14.6 &0.33&14.2 &0.39&21.3 &0.56&32.5\\
    ContentPlanner &0.53&30.4 &0.56&32.6 &0.59&35.4 &0.64&37.5\\
    \bottomrule
    \end{tabular}}
    \caption{Results on Content Planner. acc. and BLEU denote test accuracy and BLEU-2 respectively.}
    \label{tab:cpeval}
\end{table}

\subsection{Results of Content Planner}
We first report the experimental results of Content Planner. Intuitively, we use accuracy to evaluate the percentage of words that are both correct and in the right position. Following \citet{cpeval}, we also use BLEU-2 to evaluate the correctness of the words occurring in the content plan. We train Content Planner in 200 epochs with 50/100/200/500 training instances respectively and compare it with \textit{RobertaforSequenceClassification} from transformers library. The results are shown in \tabref{tab:cpeval}. We show that in the few-shot setting, Bi-LSTM+CRF performs better than fine-tuning Roberta in both word co-occurrence and positional correctness.

\subsection{Automatic Evaluation}\label{5-res}
We conduct automatic evaluations on various domains of the Wikibio dataset to prove the effectiveness of our method. We select two kinds of evaluation metrics -- \textbf{BLEU} \cite{BLEU} for evaluating overlap between the generated sentence and the gold description, and \textbf{PARENT} \cite{parent} for evaluating both the matching between the generated sentence and the reference and the fidelity of the generated sentence to the original table. Here we use F1 score of PARENT, denoted as PARENT-F.

Regarding the overlapping-based metrics BLEU, we show that our method has the best overall performance compared with all other baselines. Specifically, our approach improves 1.8\%/0.4\% BLEU score on Wiki-Humans/Wiki-Songs compared with the second best model with 200 training instances. On Wiki-Books, we improve 1.7\% BLEU score than AMG with 500 training instances. The results show that our method can produce fluent descriptions. We attribute this to the task-specific prefix that better linearizes the tabular input by comparing fine-tuning BART-large (see \secref{5-abl}).

Regarding the fidelity-based metrics PARENT, our method has better performances over AMG especially in extremely low-resource scenarios, while outperforming other baseline models. Our method performs 1.6\% PARENT-F better than AMG on average in 9 terms and 0.5\% PARENT-F worse on average in 2 terms. Reviewing their approach, AMG uses the Wikibio dataset, which is very similar to the few-shot datasets, for task adaptive pre-training. In a real-world low-resource scenario, however, it's less likely to obtain a large unlabeled corpus related to the target domain. Moreover, our approach is parameter-efficient and storage-saving. Therefore, we provide a more lightweight 
alternative with better generation fidelity than AMG.

We also implement a variant of \citet{ChenACL20}'s work with some modifications. We replace the GPT-2 with BART-large, and use prompt-tuning instead of fine-tuning to generate sentences. Therefore, the encoder in \textbf{Switch+BART} is consistent with our BART encoder in \figref{fig2}. \textbf{Switch+BART(PT)} achieves the second best performance in text fluency evaluation, obtaining the highest BLEU score on 2 terms. However, it's bad at keeping faithful to the original table when the training set size is small, which is contrary to the motivation of copy mechanism. A reasonable explanation is that the objectives of Prompt-tuning and fine-tuning are contradictory. Prompt-tuning expects that the continuous prefix can transfer to downstream tasks, while fine-tuning Pointer Generator \cite{pgen} aims to learn to copy words and to decide whether to copy or to generate. This contradiction makes Pointer Generator unable to get effective training, especially when lacking training instances. We print the selected words when the model switches to ``\textit{copy}'' state, finding that the words are far from the tokens that should be copied. We also see from \tabref{tab:parent_eval} that changing copy mechanism to the input-specific prefix significantly improves the text fidelity.

\subsection{Human Evaluation}\label{5-hum}
We randomly select 100 generated sentences (training set size is set to 500) and corresponding tables and references from the test set, then present them to three voluntary human evaluators. All volunteers are postgraduate students with extensive research experience in document summarization and natural language generation. Inspired by \citet{ChenACL20}, we assure that each sentence is evaluated according to its (1) \textit{faithfulness to the table and the reference} and (2) \textit{language fluency}. To evaluate the effectiveness of our Content Planner, we also evaluate the generated sentences according to their (3) \textit{words order correctness}. In the first task, all evaluators count the number of facts $n_{co}$ that co-occur in the table slot and the reference\footnote{Here we do not define co-occurrence as exact-matching or fuzzy-matching, instead we ask volunteers to decide co-occurrence based on human knowledge.}, and the number of facts $n_{hal}$ that contradict with/ miss from the table (i.e., hallucinated contents). The percentage of factual content is then computed through $f_p=\frac{\sum\limits_{s\in \mathcal{S}} n_{co}^s }{\sum\limits_{s\in \mathcal{S}} (n_{co}^s+n_{hal}^s) }$, where $\mathcal{S}$ denotes the select corpus. In the second task, we ask each evaluator to compare sentences in a sentence set (descriptions of an instance generated from various methods), then rank them based on their fluency and grammatical correctness. The ranking then is normalized to 0-1, the smaller the better. Finally, we average the normalized ranking of the 100 sentences to get $r_{avg}$. In the third task, all volunteers are asked to count the words order correctness. For example, given a ground-true content plan \textit{``Name Published Genre Author''} and hypothesis \textit{``A push and a shove is a 2007 novel by Christopher Kelly.''}, volunteers count the correct key pair order in the hypothesis, such as `\textit{``Published''} is in front of \textit{``Author''}. \textit{``Genre''} is not in the hypothesis, thus all its key pair order (\textit{``Name Genre''}, \textit{``Published Genre''}, \textit{``Genre Author''}) are wrong. The words order accuracy $acc_{wo}$ is averaged over all generated hypothesis. Human evaluation results are shown in \tabref{tab:humeval}. We compute the final score via $f_p-r_{avg}+acc_{wo}$ to measure the models' performance, the larger the better.

\begin{table}[t]
\centering
\resizebox{\columnwidth}{!}{
\begin{tabular}{l|ccccc}
\toprule
Model & Humans&& Books&&Songs\\
\midrule
PCG & \textbf{43.3}/\textbf{48.3} && 36.9/\textbf{46.2} && \textbf{41.7}/\textbf{45.7}\\
PCG w/o $\mathbf{c}$ & 43.2/47.2 && 37.3/44.8 && 41.5/44.5\\
PCG w/o $\mathbf{c}$\&SPA   & 41.9/46.0 && \textbf{37.4}/44.5 && 39.6/44.3\\
PCG w/o $\mathbf{p}_s$\&$\mathbf{c}$\&SPA   & 39.3/41.4 && 37.1/43.4 && 38.9/42.4\\

\bottomrule
\end{tabular}}
\caption{Ablation study results on two kinds of prefixes. $\mathbf{p}_s$, $\mathbf{c}$ and SPA denote the task-specific prefix, the input-specific prefix and Scaled Parallel Adapter respectively. In each entry, a/b denotes the BLEU/PARENT-F score.}
\label{tab:ablation}
\end{table}

\begin{table}[t]
\centering
\resizebox{\columnwidth}{!}{
\begin{tabular}{l|cccccc|c}
\toprule
Model & $f_p$&& $r_{avg}$&&$acc_{wo}$&&overall\\
\midrule
\textbf{Switch+GPT-2} & 0.62 && 0.58 && 0.71 && 0.75\\
\textbf{Prefix-tuning+T5} & 0.73 && 0.28 && 0.79 && 1.24\\
\textbf{PCG} & \textbf{0.75} && \textbf{0.20} && \textbf{0.84} && \textbf{1.39}\\

\bottomrule
\end{tabular}}
\caption{Human evaluation results.}
\label{tab:humeval}
\end{table}
\subsection{Ablation Study}\label{5-abl}
We conduct ablation studies to evaluate the improvement brought by the two kinds of prefixes we proposed. We experiment on all three datasets with the training set size of 100. The automatic results are shown in \tabref{tab:ablation}. Observing the results, we conclude that both task-specific prefix and input-specific prefix improve the fidelity of the generated sentences, while input-specific prefix contributes little to the text fluency. These conclusions are consistent with our intuitions, given that input-specific prefix aims to improve the faithfulness by planning the content. Through ablation, we show that prepending a continuous prefix to the encoder input performs better than fine-tuning BART in the few-shot scenario. In addition, adding Scaled Parallel Adapters to enhance the representation ability of prompt vectors has also proved to be effective.

\section{Conclusion}
In this paper, we propose Prompt-Controlled Generator, using two kinds of prompts to address current challenges in few-shot table-to-text generation. The task-specific prefix aims to bridge the topological structure gap between tables and sequences, which is learned via freezing the PLM and tuning the continuous prompt vectors. The input-specific prefix is designed to guide the generation process in terms of factual content and word order. We propose Content Planner to generate the input-specific prefix. Experiments on Wiki-Humans, Wiki-Books and Wiki-Songs datasets prove the effectiveness of our method from the aspects of generation fluency and text fidelity to the table.

\section*{Acknowledgements}
We thank the anonymous reviewers for their constructive comments. This work was supported by the Joint Funds of the National Natural Science Foundation of China (Grant No. U21B2020). Gongshen Liu is the corresponding author.

\bibliography{anthology}

\end{document}